\documentclass{article}
\usepackage{spconf,amsmath,graphicx}

\usepackage{hyperref}

\usepackage{xcolor}

\def\tstCommon{\texttt{tst-COMMON}}

\usepackage[
backend=biber,
style=ieee,
citestyle=numeric-comp,
maxbibnames=3,
maxcitenames=3,
doi=false,isbn=false,url=false,eprint=false
]{biblatex}

\defbibheading{bibliography}[\refname]{}

\DeclareSourcemap{
	\maps[datatype=bibtex, overwrite=true]{
		\map{
			\step[fieldsource=booktitle,
			match=\regexp{.*Interspeech.*},
			replace={Proc. Interspeech}]
			\step[fieldsource=journal,
			match=\regexp{.*INTERSPEECH.*},
			replace={Proc. Interspeech}]
			\step[fieldsource=booktitle,
			match=\regexp{.*ICASSP.*},
			replace={Proc. ICASSP}]
			\step[fieldsource=booktitle,
			match=\regexp{.*icassp_inpress.*},
			replace={Proc. ICASSP (in press)}]
			\step[fieldsource=booktitle,
			match=\regexp{.*Acoustics,.*Speech.*and.*Signal.*Processing.*},
			replace={Proc. ICASSP}]
			\step[fieldsource=booktitle,
			match=\regexp{.*International.*Conference.*on.*Learning.*Representations.*},
			replace={Proc. ICLR}]
			\step[fieldsource=booktitle,
			match=\regexp{.*International.*Conference.*on.*Computational.*Linguistics.*},
			replace={Proc. COLING}]
			\step[fieldsource=booktitle,
			match=\regexp{.*SIGdial.*Meeting.*on.*Discourse.*and.*Dialogue.*},
			replace={Proc. SIGDIAL}]
			\step[fieldsource=booktitle,
			match=\regexp{.*International.*Conference.*on.*Machine.*Learning.*},
			replace={Proc. ICML}]
			\step[fieldsource=booktitle,
			match=\regexp{.*North.*American.*Chapter.*of.*the.*Association.*for.*Computational.*Linguistics:.*Human.*Language.*Technologies.*},
			replace={Proc. NAACL}]
			\step[fieldsource=booktitle,
			match=\regexp{.*Empirical.*Methods.*in.*Natural.*Language.*Processing.*},
			replace={Proc. EMNLP}]
			\step[fieldsource=booktitle,
			match=\regexp{.*Association.*for.*Computational.*Linguistics.*},
			replace={Proc. ACL}]
			\step[fieldsource=booktitle,
			match=\regexp{.*Automatic.*Speech.*Recognition.*and.*Understanding.*},
			replace={Proc. ASRU}]
			\step[fieldsource=booktitle,
			match=\regexp{.*Spoken.*Language.*Technology.*},
			replace={Proc. SLT}]
			\step[fieldsource=booktitle,
			match=\regexp{.*Speech.*Synthesis.*Workshop.*},
			replace={Proc. SSW}]
			\step[fieldsource=booktitle,
			match=\regexp{.*workshop.*on.*speech.*synthesis.*},
			replace={Proc. SSW}]
			\step[fieldsource=booktitle,
			match=\regexp{.*Advances.*in.*neural.*information.*processing.*},
			replace={Proc. NeurIPS}]
			\step[fieldsource=booktitle,
			match=\regexp{.*Advances.*in.*Neural.*Information.*Processing.*},
			replace={Proc. NeurIPS}]
			\step[fieldsource=booktitle,
			match=\regexp{.*Workshop.*on.* Applications.* of.* Signal.*Processing.*to.*Audio.*and.*Acoustics.*},
			replace={Proc. WASPAA}]
			\step[fieldsource=booktitle,
			match=\regexp{.*International.*Spoken.*Lang.*Tran.*},
			replace={Proc. IWSLT}]
                \step[fieldsource=booktitle,
			match=\regexp{.*conf.*empirical.*natural.*language.*},
			replace={Proc. EMNLP}]
			\step[fieldsource=publisher,
			match=\regexp{.+},
			replace={{}}]
			\step[fieldsource=month,
			match=\regexp{.+},
			replace={{}}]
			\step[fieldsource=location,
			match=\regexp{.+},
			replace={{}}]
			\step[fieldsource=address,
			match=\regexp{.+},
			replace={{}}]
			\step[fieldsource=organization,
			match=\regexp{.+},
			replace={{}}]
		}
	}
}

\usepackage{booktabs}
\usepackage{multirow}

\usepackage[normalem]{ulem} 

\usepackage{booktabs}
\usepackage[plain,linesnumbered,boxed,noresetcount,noend]{algorithm2e}

\SetCommentSty{mycommfont}

\makeatletter
\newcommand{\removelatexerror}{\let\@latex@error\@gobble}
\makeatother

\usepackage[capitalize]{cleveref}

\usepackage{float}

\usepackage[normalem]{ulem}
\useunder{\uline}{\ul}{}

\usepackage{newfloat}
\DeclareFloatingEnvironment[fileext=lop]{alg}

\usepackage{silence}
\usepackage{caption}
\crefname{alg}{Algorithm}{Alg.}

\title{
Long-Form End-to-End Speech Translation via Latent Alignment Segmentation
}
\name{Peter Pol\'ak, Ond\v{r}ej Bojar
}
\address{Charles University, Czechia}
\begin{document}
%
\maketitle
\begin{abstract}

Current simultaneous speech translation models can process audio only up to a few seconds long. Contemporary datasets provide an oracle segmentation into sentences based on human-annotated transcripts and translations. However, the segmentation into sentences is not available in the real world. Current speech segmentation approaches either offer poor segmentation quality or have to trade latency for quality. In this paper, we propose a novel segmentation approach for a low-latency end-to-end speech translation. We leverage the existing speech translation encoder-decoder architecture with ST CTC and show that it can perform the segmentation task without supervision or additional parameters. To the best of our knowledge, our method is the first that allows an actual end-to-end simultaneous speech translation, as the same model is used for translation and segmentation at the same time. On a diverse set of language pairs and in- and out-of-domain data, we show that the proposed approach achieves state-of-the-art quality at no additional computational cost.

\end{abstract}

\section{Introduction}
\label{sec:intro}

Simultaneous speech translation (SST) is the task of translating speech in one language into target-language text before the speaker finishes the utterance. Traditionally, SST has relied predominantly on cascaded systems that decompose the task into multiple subtasks, including automatic speech recognition (ASR), punctuation restoration (PR), and machine translation (MT) \cite{225935, fugen2007simultaneous, bojar-etal-2021-elitr}. However, recent advancements in deep learning and the availability of abundant training data \cite{tan2018artificial, sperber-paulik-2020-speech} have led to a significant paradigm shift towards end-to-end (E2E) models. Despite the recent popularity of end-to-end SST within the research community, most research focuses on the ``short-form'' setting, which assumes that the speech input is already pre-segmented into sentences. Critically, this assumption poses an obstacle to deployment in the ``wild'', where speeches consist of several sentences ---~a~``long-form'' regime.

In the traditional cascaded approach, most segmentation methods relied on punctuation predicted by the inverse text normalization \cite{lu2010better, cho2015punctuation, cho2017nmt}. However, such an approach is impossible in the end-to-end models, as the intermediate transcript is unavailable. The E2E approach must, therefore, rely on speech-based segmentation methods. Typical choices are fixed-based segmentation, i.e., segmentation into chunks of equal length, or paused-based methods based on voice activity detection (VAD) \cite{sohn1999statistical,sinclair2014semi}. However, these segmentation approaches harm the resulting translation quality, as the translation task is sensitive to poor segmentation and generally prefers a segmentation obeying sentence boundaries \cite{sinclair2014semi}. Recent work \cite{tsiamas2022shas,fukuda2022speech} tries to predict sentence boundaries directly. However, their use in the simultaneous regime imposes further translation delay and requires additional computational resources.

This paper proposes a novel segmentation approach that leverages a popular attention-based encoder-decoder architecture with ST CTC loss \cite{deng2022blockwise,yan2022ctc}. We perform the sentence segmentation on the fly using the punctuation from the translation and speech-to-translation alignment from ST CTC. Without any external segmentation model, we show that models trained for translation only can also be used for segmentation as well. In extensive experiments on TED talks and parliamentary speeches, we show that: 

\begin{itemize}
    \item Translation models can segment speech based on the punctuation included in the translation without any special or additional training.
    \item Provided segmentation quality is equivalent to or better than the current state-of-the-art segmentation methods based on large pre-trained models.
    \item The proposed approach does not introduce any additional latency and does not need any additional computational resources.
\end{itemize}
\section{Background}
This section introduces the most essential concepts of long-form simultaneous speech translation.

\paragraph*{Incremental vs. Re-Translation SST}

SST models can be either re-translation or incremental. Re-translation models \cite{niehues16_interspeech, NiehuesPhamHa2018_1000087584} typically run their decoding every time they get a new portion of the speech. Critically, a \emph{re-translation model can revise its translation} output as more speech input is read. This design arguably makes it more difficult for the user to process the translation. On the other hand, because the model can revisit its translations, the final translation quality matches the offline translation quality.

Incremental models \cite{cho2016can,dalvi-etal-2018-incremental} differ from re-translation models in that they can only append new words to the end of the partial translation but never change the previous words. For the user, the \emph{translation changes only by incrementally getting longer}; none of the previously displayed outputs are ever modified. The incremental approach is required for certain applications (e.g., speech-to-speech translation) and can be considered easier to follow from the user's perspective.  From the long-form perspective, re-translation allows for a substantially lower latency: Imagine that punctuation prediction needs a 5-second look-ahead buffer for reliable work. In a re-translation approach, we can emit the expected translation of the 5 seconds, later fixing any punctuation mistakes. The incremental approach has to be much more conservative and delay any output until the punctuation is certain because it has no option to correct itself. In this work, \textbf{we follow the incremental approach}.\footnote{IWSLT shared tasks \cite{ansari-etal-2020-findings,anastasopoulos-etal-2021-findings,iwslt:2022,agrawal-etal-2023-findings} also follow the incremental SST approach.}

\vspace{-2mm}\paragraph*{Audio Segmentation Methods}
The simplest audio segmentation method, \textbf{fixed-length segmentation}, splits audio based on length while disregarding any information contained in the audio. More advanced strategies rely on acoustic information, typically voice activity detection (VAD). VAD concentrates solely on the presence of the speech and disregards sentence boundaries. This usually results in sub-optimal segmentation \cite{gaido2021beyond,tsiamas2022shas,fukuda2022speech} as humans place pauses inside sentences, not necessarily between them (e.g., hesitations before words with high information content, \cite{goldman1958speech}). To address this, \textbf{SHAS segmentation classifier} \cite{tsiamas2022shas} is directly trained to segment audio into sentences. The model consists of a robust pre-trained multi-lingual model XLS-R \cite{babu2021xls}, an extra Transformer \cite{vaswani2017attention} layer and a classification layer. For each speech frame, SHAS outputs the probability of whether it should be included in the segment.

To improve the quality of the VAD-based methods, \textbf{offline divide-and-conquer (DAC)} \cite{potapczyk2020srpol} and \textbf{simultaneous (SIM)} \cite{gaido2021beyond} consider the presence of speech and also the length of the resulting segments. DAC method recursively splits the audio on the longest pause until all segments are shorter than some pre-defined maximum length. The SIM method allows simultaneous segmentation (i.e., without seeing the entire recording) by segmenting on the longest pause between minimum and maximum segment length. If no pause is detected, the segmentation occurs on the maximum length. 

\vspace{-2mm}\paragraph*{Simultaneous Speech Translation Models with Latent Alignments}
\label{sec:latent}
A popular architecture for modeling speech translation is the attention-based encoder-decoder (AED) architecture. AED's advantage is the powerful cross-attention mechanism \cite{sutskever2014sequence,bahdanau2014neural} that allows the decoder to ``attend'' any portion of the source. While having overall good performance, AED models tend to hallucinate, especially in the low-latency regime \cite{iwslt:2022,polak-etal-2023-towards,polak23_interspeech}. To remedy this, an \textbf{auxiliary CTC} \cite{graves2006connectionist} directly predicting translation (ST CTC), was explored \cite{deng2022blockwise,yan2022ctc}. ST CTC provides extra regularization during training, resulting in faster and better convergence. The ST CTC output can also be used during decoding to re-score the hypotheses produced by the AED decoder \cite{watanabe2017hybrid}.\footnote{Other authors use CTC with source language transcriptions, i.e., ASR CTC. However, ASR CTC cannot be used to improve the translation quality during the inference.} We note that, unlike AED, CTC does not use cross-attention to attend the entire source speech and instead directly classifies each source-speech frame with a translation token or blank (i.e., no translation). Since each speech frame is classified with a translation or blank, this can be seen as an \textbf{explicit latent alignment between the source speech and target translation}. Any word reordering needed between the source and target languages in ST CTC happens in the encoding phase at the level of speech frames, leading to a worse quality of ST CTC alone.
\section{Method}

Our method aims to provide segmentation of the source sound by relying on the punctuation that was automatically created on the target side by the speech-to-text model. We start from ST CTC, which classifies each source speech frame with target translation, including punctuation symbols. The ST CTC output thus directly links target-side punctuation to time positions in the source. However, we must consider that the ST CTC translations are typically worse than the AED translations (e.g., \cite{yan2022ctc} report an average translation quality difference of 4 BLEU points). Also, the latent alignments of \cite{yan2022ctc} are a mere modeling tool rather than a goal product. We therefore ask two questions: \textbf{Q1: Are the latent alignments reliable?} \textbf{Q2: Are the ST CTC punctuation predictions good enough?}
To answer these questions, we propose the following two simple methods:

\vspace{-2mm}\paragraph*{Greedy Approach}
The first approach, the ``greedy'' approach, relies solely on the ST CTC predictions. For each speech frame, the greedy approach takes the translation label with the highest probability and looks if the label is a sentence punctuation symbol (i.e., ``. ! ?''). If so, the frame is labeled as a segment boundary. First, the translation of the current segment is finalized using the standard incremental beam search, and a new sentence is started. The approach is summarized in \cref{alg:greedy}.

\vspace{-2mm}\paragraph*{Align Approach}
As pointed out, the ST CTC translations are typically worse than the AED translations. Hence, the second approach, dubbed ``align'', uses the AED predictions, and the ST CTC is used only for the alignment. Specifically, the SST model provides a simultaneous translation using the standard incremental beam search. Once a sentence punctuation symbol (i.e., ``. ! ?'') is detected in the translation, we use ST CTC to find the alignment of the punctuation in the source speech. Because we assume that the incremental beam search uses CTC re-scoring that computes CTC prefix probabilities \cite{phd/de/Graves2008}, we extract the alignment as the frame with the highest CTC prefix probability, where the prefix is the generated sentence, including the sentence punctuation. This way, we obtain the alignment with one pass over the source frames. For technical details on efficient implementation of the CTC prefix probability, follow \cite{watanabe2017hybrid}. The align approach is summarized in \cref{alg:align}.

\begin{alg}[t]
    \renewcommand\figurename{Alg.}
  \removelatexerror
    \begin{algorithm*}[H]
     \footnotesize
     \DontPrintSemicolon
     \SetKwComment{Comment}{$\triangleright$\ }{}
     \SetKwInOut{Input}{Input}
     \SetKwInOut{Output}{Output}
     \Input{Streaming speech (split to small blocks), ST model (encoder, ctc, decoder)}
     \Output{Partial hypotheses}
     \For{each streaming speech block $B$}{
       $H$ $\gets$ \text{encoder}($B$) \;
       $L  \gets$ \text{ctc}($H$)          \Comment*[r]{CTC lattice; time$\times$(vocab + 1); +1 for blank} 
       \Comment*[l]{last frame $t^\text{seg}$ such that it's greedy label $v$ is a punctuation} 
       $t^\text{seg} \gets \max_t \{t \mid (\arg \max_v L_{t,v}) \in \{\text{``.''}, \text{``!''}, \text{``?''}\}\}$ \;
       \If{$t^\text{seg} \neq \emptyset$ and $t^\text{seg} \geq$ min\_len}{
        $H \gets H_{1:t^\text{seg}}$ \;
        prepend $B_{t^\text{seg}:|B|}$ to next segment \;
       }
       \Return incremental-beam-search($H, L$) \;
     }
    \end{algorithm*}
    \caption{Proposed greedy segmentation approach.}
     \label{alg:greedy}
\end{alg}

\begin{table*}[!ht]
\resizebox{\textwidth}{!}{%
\begin{tabular}{@{}ll|cccccccc|cccc@{}}
\toprule
                              &               & \multicolumn{8}{c|}{MuST-C (in-domain)}                                                                                                                                                                                                                                      & \multicolumn{4}{c}{Europarl-ST (out-of-domain)}                                                                       \\ \midrule
Type                          & Segm. method  & \multicolumn{2}{c|}{EN$\rightarrow$DE}                                 & \multicolumn{2}{c|}{EN$\rightarrow$FR}                                 & \multicolumn{2}{c|}{EN$\rightarrow$RU}                                 & \multicolumn{2}{c|}{EN$\rightarrow$ZH}            & \multicolumn{2}{c|}{EN$\rightarrow$DE}                                 & \multicolumn{2}{c}{EN$\rightarrow$FR}        \\
\multicolumn{2}{c|}{}                         & BLEU$\uparrow$      & \multicolumn{1}{c|}{LAAL$\downarrow$}            & BLEU$\uparrow$      & \multicolumn{1}{c|}{LAAL$\downarrow$}            & BLEU$\uparrow$      & \multicolumn{1}{c|}{LAAL$\downarrow$}            & BLEU$\uparrow$      & LAAL$\downarrow$            & BLEU$\uparrow$      & \multicolumn{1}{c|}{LAAL$\downarrow$}            & BLEU$\uparrow$ & LAAL$\downarrow$            \\ \midrule
                              & Oracle        & 25.4                & \multicolumn{1}{c|}{\textit{ 1750}} & 33.6                & \multicolumn{1}{c|}{\textit{ 2091}} & 16.2                & \multicolumn{1}{c|}{\textit{ 1819}} & 21.0                & \textit{ 1858} & 17.5                & \multicolumn{1}{c|}{\textit{ 2043}} & 15.8           & \textit{ 2691} \\
\multirow{-2}{*}{Offline}     & SHAS+DAC      & 24.8                & \multicolumn{1}{c|}{\textit{ 1421}} & 32.4                & \multicolumn{1}{c|}{\textit{ 2273}} & 16.0                & \multicolumn{1}{c|}{\textit{ 1466}} & 20.8                & \textit{ 1248} & 16.8                & \multicolumn{1}{c|}{\textit{ 1450}} & 15.1           & \textit{ 2177} \\ \midrule
High latency                  & SHAS+SIM-Q    & 25.0                & \multicolumn{1}{c|}{5378}                        & 33.8                & \multicolumn{1}{c|}{5733}                        & 16.0                & \multicolumn{1}{c|}{2701}                        & 20.9                & 3295                        & 16.9                & \multicolumn{1}{c|}{4833}                        & 16.4           & 5134                        \\ \midrule
                              & Fixed-length  & 22.8                & \multicolumn{1}{c|}{1339}                        & 31.3                & \multicolumn{1}{c|}{3207}                        & 14.7                & \multicolumn{1}{c|}{1418}                        & 19.6                & 1092                        & 14.0                & \multicolumn{1}{c|}{392}                         & 12.2           & 1952                        \\
                              & SHAS+SIM-L    & 23.6                & \multicolumn{1}{c|}{1582}                        & 31.3                & \multicolumn{1}{c|}{2411}                        & 15.5                & \multicolumn{1}{c|}{1687}                        & 20.4                & 1581                        & 16.1                & \multicolumn{1}{c|}{1622}                        & 14.9           & 2661                        \\
                              & Greedy (ours) & {\ul \textbf{24.2}} & \multicolumn{1}{c|}{1533}                        & {\ul \textbf{31.9}} & \multicolumn{1}{c|}{2421}                        & {\ul \textbf{16.0}} & \multicolumn{1}{c|}{1648}                        & {\ul 20.8}          & 1553                        & {\ul 16.7}          & \multicolumn{1}{c|}{1612}                        & \textbf{15.1}  & 2506                        \\
\multirow{-4}{*}{Low latency} & Align (ours)  & \dotuline{24.0}       & \multicolumn{1}{c|}{1547}                        & \dotuline{31.7}       & \multicolumn{1}{c|}{2423}                        & {\ul 15.9}          & \multicolumn{1}{c|}{1638}                        & {\ul \textbf{20.9}} & 1568                        & {\ul \textbf{16.8}} & \multicolumn{1}{c|}{1614}                        & 14.9           & 2529                        \\ \bottomrule
\end{tabular}%
}
    \caption{Systems better than the other low-latency baselines in \textbf{bold}. {\ul Underlined} and \dotuline{dotted-underlined} scores are significantly different from other low-latency baselines with {\ul $p\text{-value}<0.01$} and \dotuline{$p\text{-value}<0.05$}, respectively. Offline segmentation methods have only \textit{theoretical latency}, as the segmentation is done offline before the translation. The latency LAAL is in milliseconds.}
    \label{tab:online}
\end{table*}
\vspace{-3mm}\section{Experimental Setup}

\paragraph*{Data}
In our experiments, we use the English $\rightarrow$ German, English $\rightarrow$ French, English $\rightarrow$ Chinese, and English $\rightarrow$ Russian language pairs of the MuST-C \cite{CATTONI2021101155} data set. We use the training and validation sets during the training and tuning of the hyper-parameters for the segmentation algorithms. Finally, we use the \tstCommon{} split to report the final results. Additionally, we use the test split of Europarl-ST \cite{iranzo2020europarl} to report out-of-domain results.

\vspace{-3mm}\paragraph*{Models}
All models are attention-based encoder-decoder models. To accommodate the simultaneous regime, we adopt a blockwise encoder \cite{tsunoo2021streaming}, but any unidirectional encoder would work. We pre-process the audio with 80-dimensional filter banks. We build a unigram \cite{kudo-2018-subword} vocabulary with a size of 4000 for all language pairs. All models use a block size of 40 (1.6 s). The encoder has 12 layers, and the decoder has six layers. The model dimension is 256, and the feed-forward dimension is 2048 with four attention heads. To improve the training speed, we initialize the encoder with weights pre-trained on the ASR task of the MuST-C dataset. Further, we employ ST CTC \cite{deng2022blockwise,yan2022ctc} after the encoder with weight 0.3 during training and decoding. As a regularization, we use speed perturbation (at 0.9, 1.0, and 1.1 speeds), and to improve the long-form performance, we also include concatenation of two consecutive segments from the training data. Finally, we use checkpoint averaging for the last ten epochs. We use the ESPNet-ST toolkit \cite{yan2023espnet}

\begin{alg}[t]
  \removelatexerror
\begin{algorithm*}[H]
 \footnotesize
 \SetNoFillComment
 \DontPrintSemicolon
 \SetKwComment{Comment}{$\triangleright$\ }{}
 \SetKwInOut{Input}{Input}
 \SetKwInOut{Output}{Output}
 \Input{Streaming Speech (split to small blocks), ST model (encoder, ctc, decoder)}
 \Output{A set of partial hypotheses and scores}
 \For{each streaming speech block $B$}{
   $H$ $\gets$ \text{encoder}($B$) \;
   $L \gets$ ctc($H$)          \Comment*[r]{CTC lattice; time$\times$(vocab + 1); +1 for blank}  
   $Y \gets$ incremental-beam-search($H, L$) \;
   \Comment*[l]{index $l^\text{seg}$ of last label that is a punctuation}
   $l^\text{seg} \gets \max_l \{l \mid Y_l \in \{\text{``.''}, \text{``!''}, \text{``?''}\}\}$ \;
   
   \If{$l^\text{seg} \neq \emptyset$}{
   \Comment*[l]{time $t$ with maximal CTC prefix probability of $Y_{1:l^\text{seg}}$}
    $b^\text{seg} \gets \arg \max_t $ ctc-prefix-prob($Y_{1:l^\text{seg}}, L_{1:t}$) \;
    \If{$b^\text{seg} \geq$ min\_len}{
        $Y \gets H_{1:l^\text{seg}}$ \;
        prepend $B_{b^\text{seg} + 1:|B|}$ to next segment \;
    }
   }
   \Return Y \;
 }
\end{algorithm*}%
 \caption{Proposed alignment segmentation.}
 \label{alg:align}
\end{alg}

\vspace{-3mm}\paragraph*{Evaluation}
All models are evaluated using Simuleval \cite{ma2020simuleval} toolkit. We adopt incremental blockwise decoding \cite{tsunoo2021streaming,polak23_interspeech} with CTC incremental policy \cite{polak-etal-2023-towards}. In all our experiments, we use beam search with size 6. For the long-form evaluation, we adopt the evaluation protocol suggested by \cite{iranzo2021stream}: instead of reporting quality and latency on the document level, we align the hypothesis to the reference using a re-implementation of mwerSegmenter\footnote{For Chinese, we align on the character level instead of word level. We also tokenize the inputs before the alignment process.} \cite{matusov2005evaluating}, followed by re-segmentation into sentences based on the reference punctuation. The quality and latency metrics are then computed on the re-segmented utterances. For the translation quality, we report detokenized case-sensitive BLEU \cite{post2018call}, and for the latency, we report length-aware average lagging (LAAL) \cite{cunikit:2022,papi-etal-2022-generation}.  

\vspace{-3mm}\paragraph*{Baselines}
We use the development set to tune all hyper-parameters of the baselines. We tuned all parameters for each language pair separately. Fixed-length segmentation is tuned on interval (4, 34) seconds (s). For SHAS+DAC, we tune the maximum length between 4 and 72 s. Both proposed methods and SHAS-SIM have the minimum length between 2 and 32 s. The maximum length for SHAS+SIM was tuned relative to the minimum length on interval (1, 7) s. Because this interval influences the quality-latency tradeoff, we tune one system for latency (denoted SHAS+SIM-L) and another for quality (SHAS+SIM-Q). We found the value of approx. 2.5 s as best for SHAS+SIM-L and 7 s for SHAS+SIM-Q.

\section{Results}

We present the result in \cref{tab:online}. On in- and out-of-domain data, both proposed methods (greedy and align) outperform all low-latency baselines (fixed-length and SHAS+SIM-L) except for out-of-domain English-to-French, where the proposed align ties with SHAS+SIM-L. On average, the proposed \textbf{align approach outperforms} fixed-length by 1.6 BLEU and SHAS+SIM-L by 0.4 BLEU, and the proposed \textbf{greedy approach outperforms} fixed-length by 1.7 BLEU and SHAS+SIM-L by 0.5 BLEU across all language pairs. This answers our question Q1 --- the latent alignments are reliable for the segmentation task.  We attribute the worse quality of SHAS+SIM-L compared to the proposed methods to the SIM algorithm that forces the segmentation between minimum and maximum length. I.e., when the SHAS model does not detect any sentence boundary in this interval, SIM segments on the maximum length. In the low-latency SHAS+SIM-L, this interval is approx. 2.5 s. Considering that the average sentence length in the MuST-C test set is 5.8 s, this inevitably leads to incorrect segmentation of some sentences. On the English-to-German MuST-C test set, this occurred 203 times out of 941 segments predicted by SHAS+SIM-L in 4.7 hours, i.e., \textbf{0.6 forced sentence segmentations per minute}.

Unsurprisingly, the offline SHAS+DAC performs better than the low-latency systems. However, on average, the \textbf{proposed low-latency greedy is only 0.2 BLEU worse than the offline SHAS+DAC}. Interestingly, the high-latency SHAS+SIM-Q is better than the offline SHAS+DAC. This is probably due to the considerable delay introduced by the 7-second interval in the SHAS+SIM-Q. Since the translation model has to wait for 7 s, a large portion of each sentence is translated in an offline regime.

Counterintuitively, the \textbf{greedy approach outperforms the align approach} slightly (only 0.1-0.2 BLEU). Because the CTC translation quality is worse than that of the AED \cite{yan2022ctc}, we would expect the align approach to reach a better quality. A possible answer might be a mismatch between the ST CTC and AED predictions that leads to a slightly poorer alignment. This answers our question Q2 --- the ST CTC punctuation predictions are suitable for the segmentation.

\begin{table}[t]
\centering
\resizebox{\columnwidth}{!}{%
\begin{tabular}{@{}l|ccccc@{}}
\toprule
Segm. method  & Segm. param.$\downarrow$ & Total param.$\downarrow$ & RTF$\downarrow$ & LAAL$\downarrow$ & BLEU$\uparrow$ \\ \midrule
Fixed-length  & \textbf{0}                & \textbf{45 M}             & 0.46            & \textbf{1339}    & 22.8           \\
SHAS+SIM-L    & 208 M                     & 253 M                     & 0.61            & 1582             & 23.6           \\
Greedy (ours) & \textbf{0}                & \textbf{45 M}             & 0.42            & 1533             & \textbf{24.2}  \\
Align (ours)  & \textbf{0}                & \textbf{45 M}             & \textbf{0.41}   & 1547             & 24.0           \\ \bottomrule
\end{tabular}%
}
\caption{Performance comparison of low latency segmentation methods on English-to-German MuST-C test set. Real-time factor (RTF) measured on Intel i7-10700 using a single thread. Better values in \textbf{bold}. Total param. is the total number of parameters, including the translation model.}
\label{tab:performance}
\end{table}

In \cref{tab:performance}, we compare the computational complexity of the low latency systems. The proposed segmentation methods, like the fixed-length method, \textbf{do not introduce new segmentation parameters}. The proposed methods have about \textbf{30~\% lower real-time factor} (RTF) than the SHAS+SIM-L, as they do not have to evaluate the additional segmentation model. Interestingly, the fixed-length method has a slightly higher RTF. The probable cause is the quadratic complexity of the AED decoder and the length of an average segment proposed by the segmentation methods: the fixed-length method uses 20 s (was found to maximize the translation quality on the development set) and the proposed align method produces segments of an average length of 8.5~s. 
\section{Conclusion}
In this paper, we presented two simple speech segmentation methods introducing new state-of-the-art performance to simultaneous speech segmentation. A thorough evaluation on in- and out-of-domain data shows that the proposed methods offer the best quality with the same latency and have the smallest computational footprint. To the best of our knowledge, our methods are the first that allow an actual end-to-end simultaneous speech translation, as they use the translation model for the joint translation and segmentation without explicitly modeling the segmentation. In future research, we will explore the properties of latent alignments, including latent alignments from other architectures.

\vfill\pagebreak


\section{REFERENCES}
{
\printbibliography
}

\end{document}